\documentclass[letterpaper, 10 pt, conference]{ieeeconf}  

\IEEEoverridecommandlockouts                              
\makeatletter

\newcommand{\Rmnum}[1]{\expandafter\@slowromancap\romannumeral #1@}
\makeatother

\usepackage{amsfonts}
\usepackage{multicol} 
\usepackage{cite}
\usepackage{graphics} 
\usepackage{epsfig} 
\usepackage{times} 
\usepackage{booktabs}
\usepackage{float}
\usepackage{color}
\usepackage{comment}
\usepackage{amsmath} 
\usepackage{amssymb}  
\usepackage{mathptmx} 
\usepackage{mathrsfs}
\usepackage{sansmath}
\usepackage{algpseudocode}
\usepackage{algorithmicx,algorithm}
\usepackage{cases}
\usepackage{newtxtext} 
\usepackage{textcomp} 
\usepackage[varg,cmintegrals,cmbraces]{newtxmath}
\usepackage{bm} 

\title{\LARGE \bf
 Deep Reinforcement Learning Based Tracking Control of an Autonomous Surface Vessel in Natural Waters
}
\author{
Wei Wang$^{1,2,\ast}$, Xiaojing Cao$^{3,\ast}$, Alejandro Gonzalez-Garcia$^{2,4}$, Lianhao Yin$^{1}$, Niklas Hagemann$^{2}$,  \\ Yuanyuan Qiao$^{2,3}$,  Carlo Ratti$^{2}$, and Daniela Rus$^{1}$
\thanks{This work was supported by grant from the Amsterdam Institute for Advanced Metropolitan Solutions (AMS) in Netherlands. L. Yin was supported by Knut and Alice Wallenberg Foundation.}
\thanks{$^{1}$W. Wang, L. Yin, and D. Rus are with the Computer Science and Artificial Intelligence Lab (CSAIL),  Massachusetts Institute of Technology, Cambridge, MA 02139 USA. }%
\thanks{$^{2}$W. Wang, A. Gonzalez-Garcia, N. Hagemann, Y. Qiao, and C. Ratti  are with the SENSEable City Laboratory, Massachusetts Institute of Technology, Cambridge, MA 02139 USA.}
\thanks{$^{3}$X. Cao, and Y. Qiao are with Intelligent Perception and Computing Research Center, the School of Artificial Intelligence, Beijing University of Posts and Telecommunications (BUPT), Beijing 100876, China.}
\thanks{$^{4}$A. Gonzalez-Garcia is with MECO Research Team, Department of Mechanical Engineering, KU Leuven, Belgium and Flanders Make@KU Leuven, Belgium.}
\thanks{$^{\ast}$These authors contribute equally to this work.}
}
\begin{document}

\maketitle
\thispagestyle{empty}
\pagestyle{empty}

\begin{abstract}
Accurate control of autonomous marine robots still poses challenges due to the complex dynamics of the environment. In this paper, we propose a Deep Reinforcement Learning (DRL) approach to train a controller for autonomous surface vessel (ASV) trajectory tracking and compare its performance with an advanced nonlinear model predictive controller (NMPC) in real environments. Taking into account environmental disturbances (e.g., wind, waves, and currents), noisy measurements, and non-ideal actuators presented in the physical ASV, several effective reward functions for DRL tracking control policies are carefully designed. The control policies were trained in a simulation environment with diverse tracking trajectories and disturbances. The performance of the DRL controller has been verified and compared with the NMPC in both simulations with model-based environmental disturbances  and in natural waters. Simulations show that the DRL controller has 53.33\% lower tracking error than that of NMPC. Experimental results further show that, compared to NMPC, the DRL controller has 35.51\% lower tracking error, indicating that DRL controllers offer better disturbance rejection in river environments than NMPC. 
\end{abstract}

\section{Introduction}

Autonomous Surface Vessels (ASVs) deliver solutions for a variety of applications, including environmental monitoring, security patrols, search and rescue, bridge and other infrastructure inspections, and military and naval operations \cite {liu2016unmanned, doniec2010complete, Li9341105,Corke2007a, Doniec1648, Leonard1639838, paull2014auv, Dhariwal4399056, AzzeriaAdnanb1727,li2022aerial,yan2022real}.
Trajectory tracking is an essential capability for ASVs because many other higher-level robot tasks such as navigation, docking and multi-robot coordination require time-critical tracking.

Trajectory tracking is a difficult task because ASVs usually follow a predefined trajectory while being exposed to environmental disturbances caused by waves, winds, and ocean currents \cite{KARIMI2021}.
Several studies have addressed the ASV trajectory tracking control including the sliding mode method \cite{Wang2019a}, input-output feedback linearization \cite{Paliotta2019} and adaptive control \cite{VAN2019}. 
In recent years, Model Predictive Control (MPC) has been proposed to solve the trajectory tracking control problem \cite{7231629} and proven to be the most efficient controller for ASV trajectory tracking ~\cite{WeiICRA2018,WeiIROS2019,WeiIROS2020a, WangDistributedControl2020IROS,WangICRA2021}. However, MPC is susceptible to model uncertainty and external perturbations.
While robust controllers can in some respect deal with time-varying disturbances and uncertainties, designing robust controllers for ASVs is still difficult. For example, robust controllers can cause instability with improper system models or lack of tuning.

Model-free Reinforcement Learning (RL) trains a control policy based on the interaction of agents with their environment\cite{Sutton98} and does not require an accurate model. Therefore, RL has been proposed for ASV control problems \cite{Martinsen2020ReinforcementLT,Zhang9454561,Wang9154585,WANG202226}. However, only \cite{Martinsen2020ReinforcementLT} provides an experimental evaluation of an RL approach, introducing sea trials for trajectory tracking of a fully-actuated ASV. Besides, the approach is model-based, learning model parameters, not only the control policy.



Moreover, Deep RL (DRL), learning from more complex and dynamic environments and scenarios \cite{zhao2020path}, has led to a few interesting developments in the learning of control laws for ASVs \cite{Zhang9802680,zhao2020path, CHENG2018, woo2019deep,ZHOU2022}.
The Deep Q-Network (DQN) is a powerful actor-critic DRL algorithm, and it has been applied for ASV path following in \cite{zhao2020path}, and for obstacle avoidance in \cite{CHENG2018}. Nonetheless, DQN can only deal with discrete and low-dimensional action spaces. Hence, the Deep Deterministic Policy Gradient (DDPG) algorithm was developed to benefit from the actor-critic architecture, and to deal with continuous action spaces \cite{lillicrap2015continuous}. DDPG was then implemented in \cite{woo2019deep} for ASV path following control, presenting simulation and experimental results. Likewise, DDPG was used for obstacle avoidance in \cite{ZHOU2022}, illustrating an improvement when compared against DQN-based strategies.

Nevertheless, the vast majority of DRL controllers for ASVs are limited to simulation, as experimental evaluations are scarce.
Furthermore, current DRL methods are often used to solve the path following problem. The trajectory tracking problem has rarely been tackled in the ASV literature, partly because it requires more control constraints that require a complicated multi-action joint policy in the DRL control system. Additionally, the performance of the current ASV DRL controller is usually not compared to traditional controllers, especially against techniques such as MPC.

Based on the above discussion, this paper focuses on solving the trajectory-tracking problem for a physical ASV using DRL. Physical robots typically exhibit complex behaviors in real environments in the presence of environmental disturbances, noisy measurements, and non-ideal actuators. We propose a framework for creating an efficient DRL reward function that takes into account these complex behaviors. 

The main contributions of our work are summarized as follows:\\
$\tiny {\bullet}$ A DRL controller for trajectory tracking of real ASVs;\\
$\tiny {\bullet}$ Numerical simulations considering model-based environmental disturbances which illustrate the potential of the DRL policy;\\
$\tiny {\bullet}$ Extensive natural-water experiments verifying the effectiveness of DRL controllers;\\
$\tiny {\bullet}$ Comparisons between DRL and MPC showing that: 1) DRL outperforms MPC for tracking error; 2) DRL does better than MPC at rejecting disturbances.  

\section{ASV Overview and Its Dynamic Model}
In this section, we introduce the ASV prototype and its simplified mathematical model. The simplified model serves as the basis for the simulation environment of the proposed DRL controller in the next section.

\subsection{Surface Vessel}
\begin{figure}[htb]
    \centering
    \includegraphics[width=0.8\linewidth] {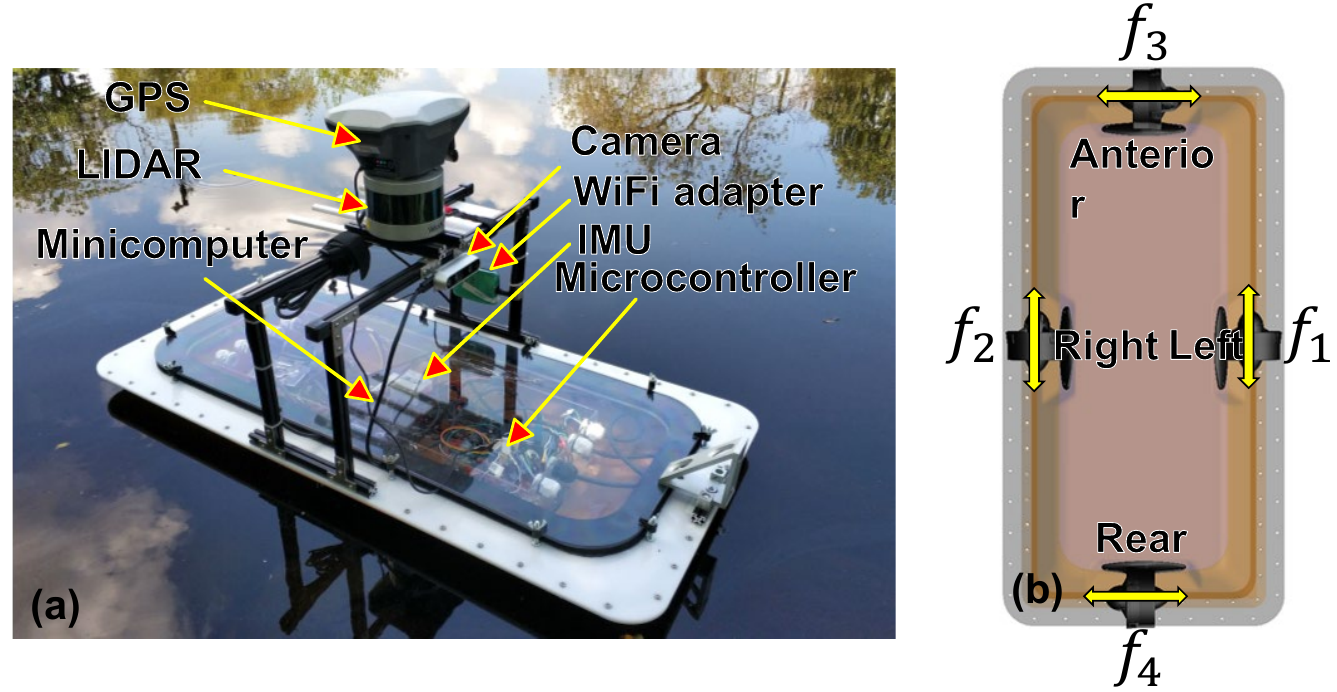}
    \caption{The surface vessel. (a) The prototype (side view); (b) thruster configuration (bottom view). $f_1$, $f_2$, $f_3$ and $f_4$ denote the forces generated by left, right, anterior and rear thrusters, respectively.}
    \label{Roboat}
\end{figure}

The vessel has four thrusters around the hull to achieve holonomic motions as shown in Fig. \ref{Roboat}.
An Intel NUC is adopted as the main controller running the Robotic Operating System (ROS). Moreover, an auxiliary microprocessor (STM32F103) is used for real-time actuator control. The robot has multiple onboard sensors including a 3D LiDAR (Velodyne, Puck VLP-16), an IMU, a camera, and a GPS sensor. The camera and GPS are not used in this paper.
The vessel weighs around 15 kg, and its dimensions are 0.90 $\times$  0.45 $\times$ 0.15 m.  It is powered by an 11.1 V Li-Po battery which lasts around three hours. More details of the robot hardware can be found in \cite{WeiICRA2018}.

\subsection{Dynamic Model of ASV}
The dynamics of our vessel is described by the following nonlinear differential equation \cite{WeiICRA2018}
\begin{eqnarray}\label{MFCPredynamics}
&&\dot{\bm{\eta}}=\mathbf{T}(\bm{\eta})\mathbf{v}  \label{MFCPredynamicsA}\\
&&\dot{\mathbf{v}}=\mathbf{M}^{-1}(\bm{\tau}+\bm{\tau}_{\text{env}})-\mathbf{M}^{-1}(\mathbf{C}(\mathbf{v})+\mathbf{D}(\mathbf{v}))\mathbf{v} \label{MFCPredynamicsB}
\end{eqnarray}
where  $\bm{\eta}=[x \quad y \quad \psi]^{T} \in\mathbb{R}^{3}$ is the position and heading angle of the vessel in the inertial frame; $\mathbf{v}=[u \quad v \quad w]^{T}\in \mathbb{R}^{3}$ denotes the vessel velocity, which contains the surge velocity ($u$), sway velocity ($v$), and yaw rate ($w$) in the body fixed frame; $\mathbf{T}(\bm{\eta})\in \mathbb{R}^{3\times3}$ is the transformation matrix converting a state vector from body frame to inertial frame; $\mathbf{M} \in \mathbb{R}^{3\times3}$ is the positive-definite symmetric added mass and inertia matrix; $\mathbf{C}(\mathbf{v})\in\mathbb{R}^{3\times3}$ is the skew-symmetric vehicle matrix of Coriolis and centripetal terms; $\bm{\tau}_{\text{env}} \in\mathbb{R}^{3}$ is the environmental disturbances from the wind, currents and waves; $\mathbf{D}(\mathbf{v})\in\mathbb{R}^{3\times3}$ is the positive-semi-definite drag matrix-valued function; $\bm{\tau}=[\tau_u~ \tau_v~ \tau_w]^{T} \in\mathbb{R}^{3}$ is the force and torque applied to the vessel in all three DOFs, and is defined as follows
\begin{eqnarray}\label{AppliedForceMaxtrix}
\bm{\tau}
=\mathbf{B}\mathbf{u}
=
\left[
 \begin{array}{cccc}
1                      &  1                                      &    0                       & 0\\
0                      &  0                                      &    1                       & 1\\
\dfrac{a}{2}&-\dfrac{a}{2}                 &     \dfrac{b}{2} &-\dfrac{b}{2}
\end{array}
\right]
 \left(
 \begin{array}{c}
f_1\\
f_2\\
f_3\\
f_4
\end{array}
\right)
\end{eqnarray}
where $\mathbf{B}\in\mathbb{R}^{4\times3}$ is the control matrix describing the thruster configuration and $\mathbf{u}=[f_1 \quad f_2 \quad f_3 \quad f_4]^{T}\in\mathbb{R}^{4}$ is the control vector where $f_1$, $f_2$, $f_3$ and $f_4$ represent the left, right, anterior, and rear thrusters, respectively; $a$ is the distance between the transverse thrusters and $b$ is the distance between the longitudinal thrusters.
We  refer  the  readers  to~\cite{WeiICRA2018} for more details of $\mathbf{M}$, $\mathbf{C}(\mathbf{v})$ and  $\mathbf{D}(\mathbf{v})$.

By combining (\ref{MFCPredynamicsA}) and (\ref{MFCPredynamicsB}), the complete dynamic model of the vessel is further reformulated as follows
\begin{eqnarray}\label{MPCdynamics}
\dot{\mathbf{q}}(t)=f(\mathbf{q}(t),\mathbf{u}(t))
\end{eqnarray}
where $\mathbf{q}=[x \quad y  \quad \psi \quad  u \quad  v \quad  w]^{T}\in \mathbb{R}^{6}$  is the state vector of the vessel, and $f(\cdot, \cdot): \mathbb{R}^{n_q}\times \mathbb{R}^{n_u} \longrightarrow \mathbb{R}^{n_q}$ denote the continuously differentiable state update function. The system model describes how the full state $\mathbf{q}$  changes in response to applied control input $\mathbf{u} \in\mathbb{R}^{4}$.
(\ref{MPCdynamics}) will act as a simulator in the simulation environment to train or test the control policy. 

\section{DRL Framework for ASV Trajectory Tracking}
In this section, we propose a DRL framework to solve the trajectory tracking problem of a physical ASV. 
\subsection{Problem Forumlation}
Let $\bm{\eta}_{\text{d}}(t)=[x_{\text{d}}(t) \quad  y_{\text{d}}(t) \quad  \psi_{\text{d}}(t)]^{T} \rightarrow \mathbb{R}^{3}$ be a given sufficiently smooth time-varying desired trajectory with bounded time-derivatives. The trajectory tracking problem is: 
Design a controller such that all the closed-loop signals are bounded and the tracking error $\|\bm{\eta}(t)-\bm{\eta}_{\text{d}}(t)\|$ converges to a neighborhood of the origin that can be made arbitrarily small. 

\subsection{Method}
We can formulate the trajectory tracking task as a discrete-time optimization problem and use DRL to solve this control task in which robots learn based on trial and error to modify their actions to maximize the gains from interaction with their environment \cite{sutton2018reinforcement}.
In this paper, for verifying our proposed learning framework, we select DDPG network \cite{lillicrap2015continuous} as a DRL benchmark for ASV trajectory tracking due to its capability of continuous control and previous success in ASV control problems \cite{woo2019deep,ZHOU2022}. Each actor and critic network has two fully connected hidden layers with 300 neurons. The activation function between the hidden layers is Relu, and that of the output layer is Tanh.
\subsection{Control Architecture}
\begin{figure}[htb]
        \centering
        \includegraphics[width=0.9\linewidth] {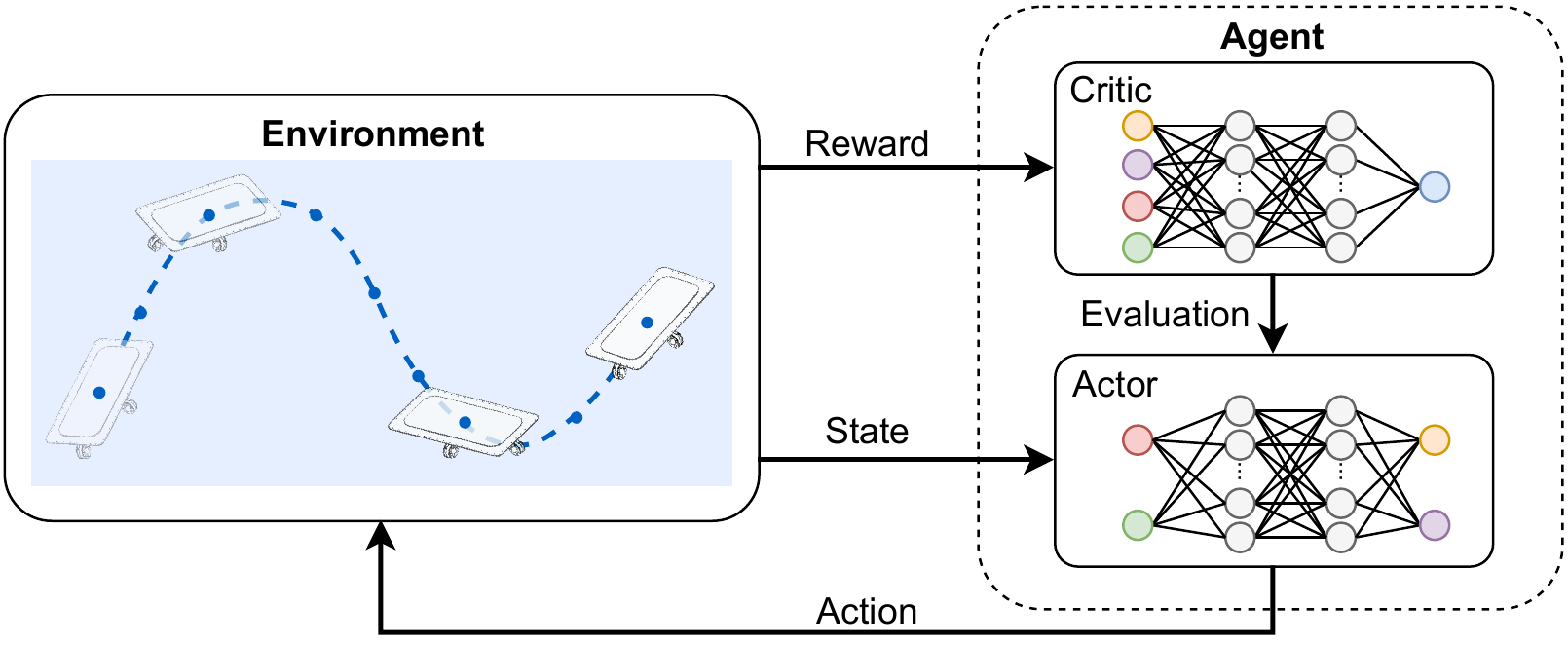}
        \caption{Structure of the DRL controller.}
        \label{DRLControlSystem}
\end{figure}

Our goal is to learn a policy $\pi_{\theta}(\cdot)$ that takes robot state $\mathbf{q}$ and reference commands $\bm{\eta}_{\text{d}}(t)$ as input and gives as output force commands $\mathbf{u}$, which are further converted into Pulse-Width Modulation (PWM) signals to drive the thrusters.
Our DRL framework is illustrated in Fig. \ref{DRLControlSystem}. 

\begin{figure}[htb]
        \centering
        \includegraphics[width=0.7\linewidth] {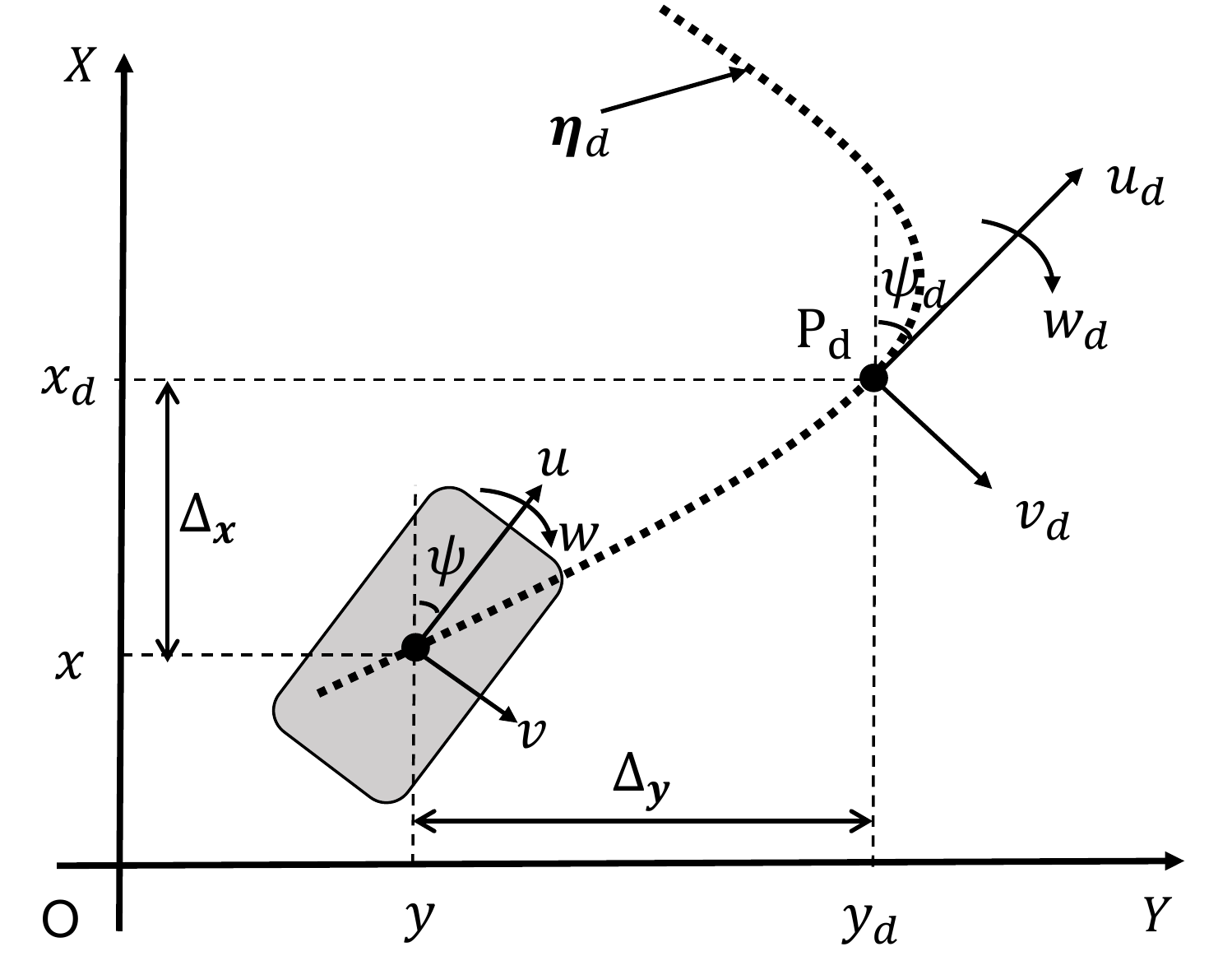}
        \caption{Representation of a surface vessel in DRL trajectory tracking.}
        \label{StateFrame}
\end{figure}

\noindent \textbf{Observation Space:}
The policy $\pi_{\theta}(\cdot)$ takes as input a history of previous observations (see in Fig. \ref{StateFrame}) and actions denoted by $\mathbf{s}_{{t-H}:t}$ where $\mathbf{s}_t=[\bm{q}_t, \bm{q}^{\text{d}}_t,\mathbf{a}_{t-1}]$ is the observation.
 $\mathbf{q}_t=[x(t) \quad y(t)  \quad \psi(t) \quad  u(t) \quad  v(t) \quad  w(t)]^{T}\in \mathbb{R}^{6}$ is robot state at time $t$ defined in Section II, $\bm{q}^{\text{d}}_t=[x_{\text{d}}(t) \quad y_{\text{d}}(t)  \quad \psi_{\text{d}}(t) \quad  u_{\text{d}}(t) \quad  v_{\text{d}}(t) \quad  w_{\text{d}}(t)]^{T}\in \mathbb{R}^{6}$ is the reference state for tracking and $\mathbf{a}_{t-1}$ is the action at the last time step $t-1$. 

\noindent \textbf{Action Space:}
The action, $\mathbf{a}_t \in \mathbb{R}^{4}$ is defined as $\mathbf{a}_t = \mathbf{u}_t$ where $\mathbf{u}_t=[f_1(t) \quad f_2(t) \quad f_3(t) \quad f_4(t)]^{T}$ is expressed in (\ref{AppliedForceMaxtrix}).

\noindent \textbf{Reward Function:}
Reward function follows with tracking task reward terms as well as a set of auxiliary terms for stability (velocity variation penalties), smoothness (action variation penalties), and minimal energy consumption. 

We define Euclidean distance error as $e_{\text{p}}=[(x-x_{\text{d}})^2+(y-y_{\text{d}})^2]^{1/2}$.Then, $r_{\text{p}}$ is designed to minimize the position tracking error, denoted as follows:
\begin{eqnarray}\label{r_L}
r_{\text{p}}=2^{-k_1(\mu\|e_{\text{v}}\|+1)e_{\text{p}}}-1
\end{eqnarray}
where $\mu$ and $k_1$ are constant parameters, and $e_{\text{v}}=[(u-u_{\text{d}})^2+(v-v_{\text{d}})^2]^{1/2}$. 
$e_{\text{v}}$ is used to smooth linear velocity variations. 

\begin{figure}[htb]
        \centering
        \includegraphics[width=0.75\linewidth] {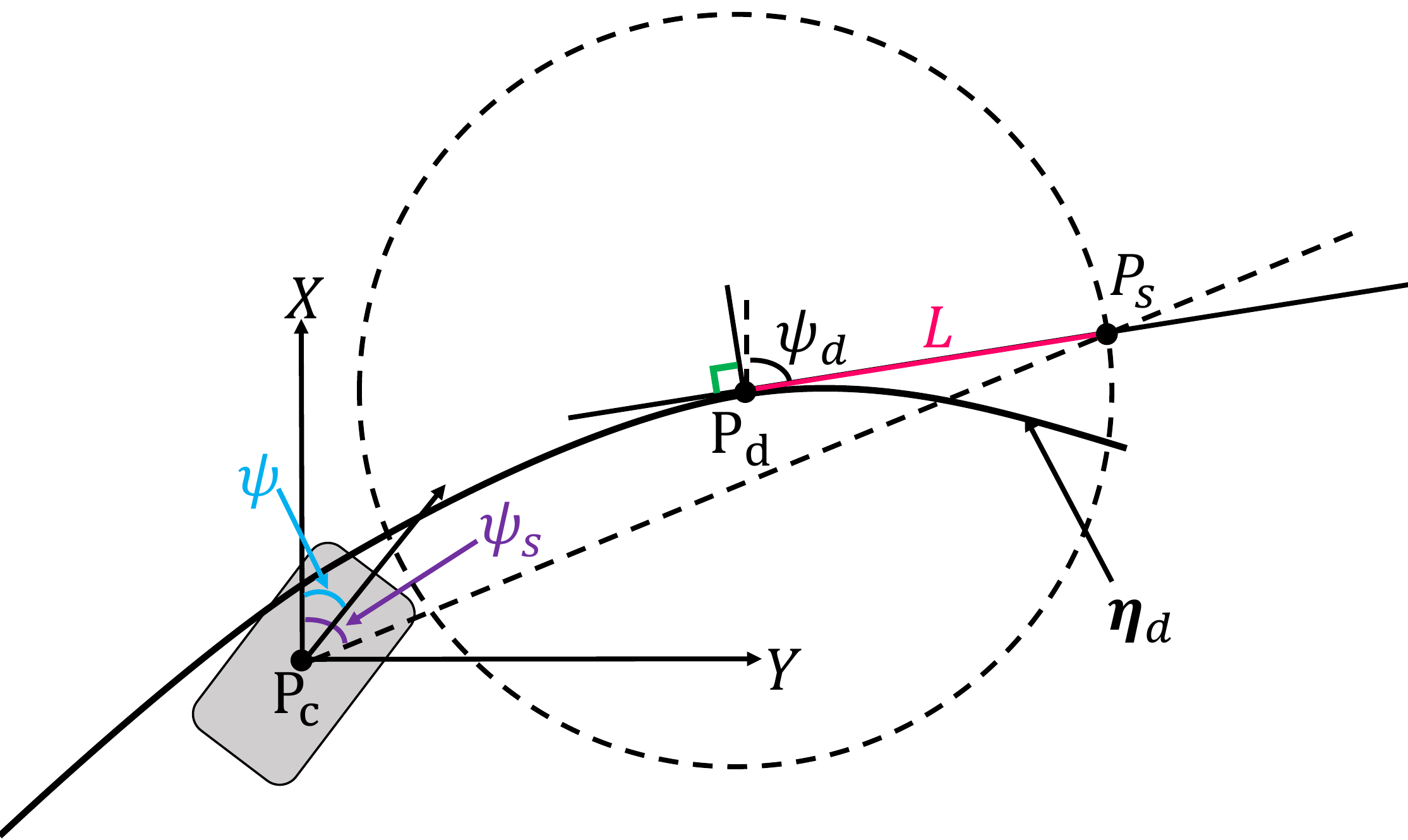}
        \caption{LOS guidance geometry for curved trajectory.}
        \label{LOSFrame}
\end{figure}

Second, $r_{\psi}$ is designed to minimize the heading error, expressed as follows:
\begin{eqnarray}\label{e_{psi}}
r_{\psi}=
\left\{\begin{array}{ll}
                    e^{-k_2\|(\psi-\psi_{\text{s}})\|},          &  \mbox{if}~\|\psi-\psi_{\text{s}}\| \le 0.5\pi\\
                    -e^{k_2(\|\psi-\psi_{\text{s}}\|-\pi)},      &   \mbox{if}~\|\psi-\psi_{\text{s}}\| > 0.5\pi \\
                 \end{array} 
                 \right.,
\end{eqnarray}
where $k_2$ is a constant,  $e_{\psi} =\psi-\psi_{\text{d}}$ is the heading error, and $\psi_{\text{s}}$ is the sight heading which replaces the reference heading. 
The sight heading is calculated based on a line of sight (LOS) guidance method \cite{fossen2003line}, shown in Fig. \ref{LOSFrame}. 
We define $P_{\text{c}}(x,y)$,  $P_{\text{d}}(x_{\text{d}},y_{\text{d}})$ and  $P_{\text{s}}(x_{\text{s}},y_{\text{s}})$ are the current position, desired position and sight position of robot, respectively. The sight position $P_{\text{s}}(x_{\text{s}},y_{\text{s}})$ is $L$ length forward along the tangent line at the reference position $P_{\text{d}}(x_{\text{d}},y_{\text{d}})$. Then the sight heading $\psi_{\text{s}}$ is computed as the angle between the vector from $P_{\text{c}}(x,y)$ to $P_{\text{s}}(x_{\text{s}},y_{\text{s}})$ and $X$ axis of the inertial frame.

Moreover, reward term $r_{w}$ is designed to punish variations in angular velocity, denoted as: 
\begin{eqnarray}\label{e_{w}}
r_{w}= e^{-k_3 \|e_w\|}-1
\end{eqnarray}
where $k_3$ is a constant and $e_w=w-w_{\text{d}}$.

Furthermore, reward term $r_{\text{a}}$ is proposed to suppress variations in actions, expressed as follows: 
\begin{eqnarray}\label{ra}
r_{\text{a}}=e^{-k_4\Delta_{\text{a}}}-1
\end{eqnarray}
where $k_4$ and is a constant, $\Delta_{\text{a}}(t) =\sum_{i=1}^{4}(f_i(t) -f_i(t-1))$.


Finally, a reward term is mathematically defined to minimize the energy consumption of the propulsion system:
\begin{eqnarray}\label{re}
r_{\text{e}}=e^{-k_5 E_{\text{p}}}-1
\end{eqnarray}
where $k_5$ is a constant, and $E_{\text{p}} = \sum_{i=1}^{4}f_i^2$ represents the energy consumption of the thrusters.This reward term always punishes high energy costs-related activities, which leads to the convergence of control policies to minimal energy policies.

In summary, the whole reward function is formulated as follows
\begin{eqnarray}\label{r_{total}}
r=\left\{\!\!\!\begin{array} {ll}
            \lambda_1r_{\text{p}}+\lambda_2r_{\psi}+\lambda_3r_{w}+\lambda_4r_{\text{a}}+\lambda_5r_{\text{e}},   &\!\!\!\!\! \mbox{if}~e_{\text{p}} \le e_{\text{bound}}\\
            -25,                                                                           &\!\!\!\!\!  \mbox{if}~  e_{\text{p}}>e_{\text{bound}}
            \end{array}\right.\!\!\!\!\!, 
\end{eqnarray}
where $e_{\text{bound}}$ is a boundary distance, $\lambda_1,\lambda_2,\lambda_3,\lambda_4,\lambda_5$ are weight coefficients for the five reward terms.
If the vessel sails out of the boundary error, the total reward will turn into large punishment.
\subsection{Simulation Environment}
Computational fluid dynamics is a sufficiently accurate simulation environment but it is still too computationally expensive to be adopted during the training procedure.
To balance accuracy and computing speed, we construct an ASV simulator using the differentiable nonlinear equation (\ref{MPCdynamics}) defined in Section II. The unknown hydrodynamic parameters $\mathbf{M}$, $\mathbf{C}$, and $\mathbf{D}$ were identified by a nonlinear least squares method based on Trust Zone Reflection Algorithm \cite{WeiICRA2018}. The identified parameters for our vessel are found as $m_{11} = 12$ kg, $m_{22} = 24$ kg, $m_{33} = 1.5$ kg\,m$^{2}$, $d_{11} = 6$ kg\,s$^{-1}$, $d_{22} = 8$ kg\,s$^{-1}$, and $d_{33} = 1.35$ kg\,m$^2$\,s$^{-1}$. At each time step, the simulator updates the vessel's state once it receives a new control action from the control policy. 
\subsection{Training}

We train a control policy in simulation and transfer it
to the real world without fine-tuning. The training contains random trajectory, initial position and measurement disturbance. 
A diversity of sinusoidal waves are selected as the training trajectories because they generate a coupled motion in the surge, sway, and yaw degrees of freedom, at time-varying velocities. 
The robot is randomly located near the initial target position, at the start of each episode. Meanwhile, the robot heading is also randomly generated.
The episode will stop if 1) the tracking task fails which means that the vessel sails out of the boundary, i.e., $e_{\text{p}}>e_{\text{bound}}$ where $e_{\text{bound}}= 1$ m; 2) The runtime comes to its maximum value, which is set to $t_{\text{max}} = 30$ s.

The measured state $\mathbf{q}$ of the robot are corrupted by zero-mean
Gaussian noise $N(0, 0.1I_6)$, where $I_6$ is a 6D identity matrix.  Moreover, the parameter values in the reward function (\ref{r_{total}}) are listed in Table \ref{Parametervaluesinrewardfunction}.
\begin{table}[ht]
\small
\begin{center}
\caption{Parameter values in the reward function}
\setlength{\tabcolsep}{2.5mm}
\label{Parametervaluesinrewardfunction}
\begin{tabular}{cccccccc}
\toprule[0.9pt]
Parameters               &$\mu$        & $k_1$       & $k_2$       & $k_3$       &  $k_4$       & $k_5$ \\
Values                   &3            &  4          & 3           & 1           & 0.033        & 0.017 \\
\toprule[0.3pt]
Parameters               & $\lambda_1$ & $\lambda_2$ & $\lambda_3$ & $\lambda_4$ & $\lambda_5$  &  $e_{\text{bound}}$\\
Values                   & 1.5         & 0.5         & 1           & 0.5         & 0.2          & 1 m \\
\bottomrule[0.9pt]
\end{tabular}
\end{center}
\end{table}
We add an Ornstein-Uhlenbeck process \cite{uhlenbeck1930theory} onto the action to improve the exploration with $\theta = 0.2$ and $\sigma = 0.15$. Policy weights were selected after around 3000 training episodes. 

\section{Simulations and Experiments}
In this section, extensive simulations and experiments are conducted with the surface vessel shown in Fig. \ref{Roboat} to validate the tracking capabilities of the developed DRL control strategy. Furthermore, we systematically compare the performance of the proposed DRL controller with NMPC in terms of tracking error, energy consumption, and force allocation strategy, and finally obtain several interesting capabilities of the DRL controller. 
Both simulation and experiments are executed on a small-form-factor barebone computer, Intel\textsuperscript{\textregistered} NUC (NUC7i7DNH), which runs a robotics middleware, Robot Operating System (ROS). 

\subsection{Simulations}
We use root mean square error (RMSE) to represent tracking error. 
In general, thruster power consumption is positively correlated to its thrust. Hence, the mean power consumption $E_{\text{ave}}$ of the controller is represented as follows:
\begin{eqnarray}\label{averagenergy}
E_{\text{ave}} \propto \sqrt{\dfrac{\sum_{i=1}^{N}(|f_1^{i}|+|f_2^{i}|+|f_3^{i}|+|f_4^{i}|)}{N}} 
\end{eqnarray}
where $N$ is the number of the data points.

\subsubsection{Ablation Study}
Typical reward functions for ASV tracking only consider the terms (\ref{r_L}) and (\ref{e_{psi}}) to minimize the position and heading errors. By contrast, our DRL framework contains three more reward terms which largely guarantees that the trained control policy in simulation can be transferred to the real world without fine-tuning.
An ablation study was conducted in simulation to demonstrate the benefits of the proposed reward function design in our DRL framework.

\begin{figure}[htb]
        \centering
        \includegraphics[width=0.9\linewidth] {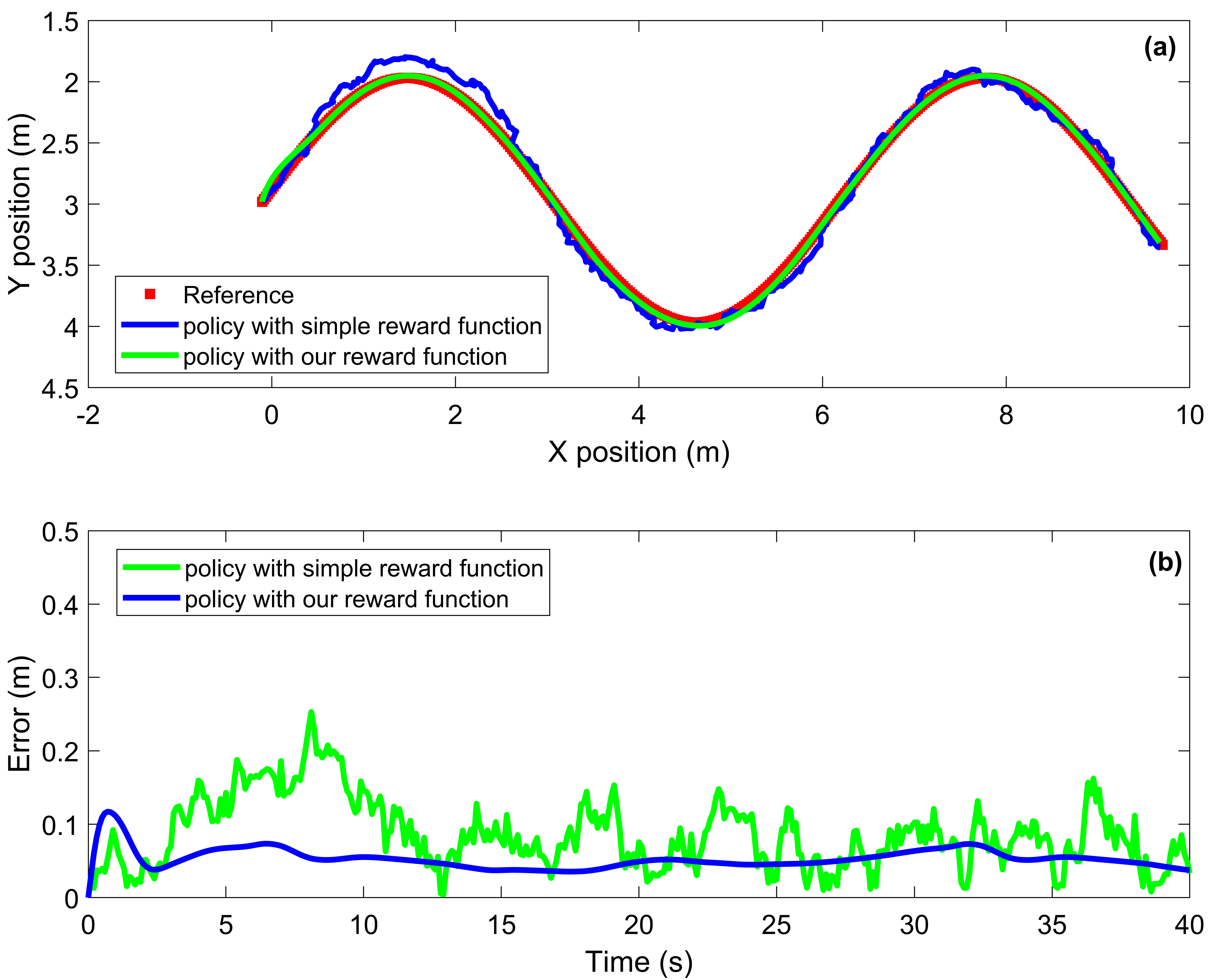}
        \caption{Comparison of sinusoidal tracking performance in DRL using a simple reward function and our reward function. (a) Trajectory and (b) tracking error.}
        \label{DRLbadrewardvsDRLreward}
\end{figure}

\begin{figure}[htb]
        \centering
        \includegraphics[width=0.9\linewidth] {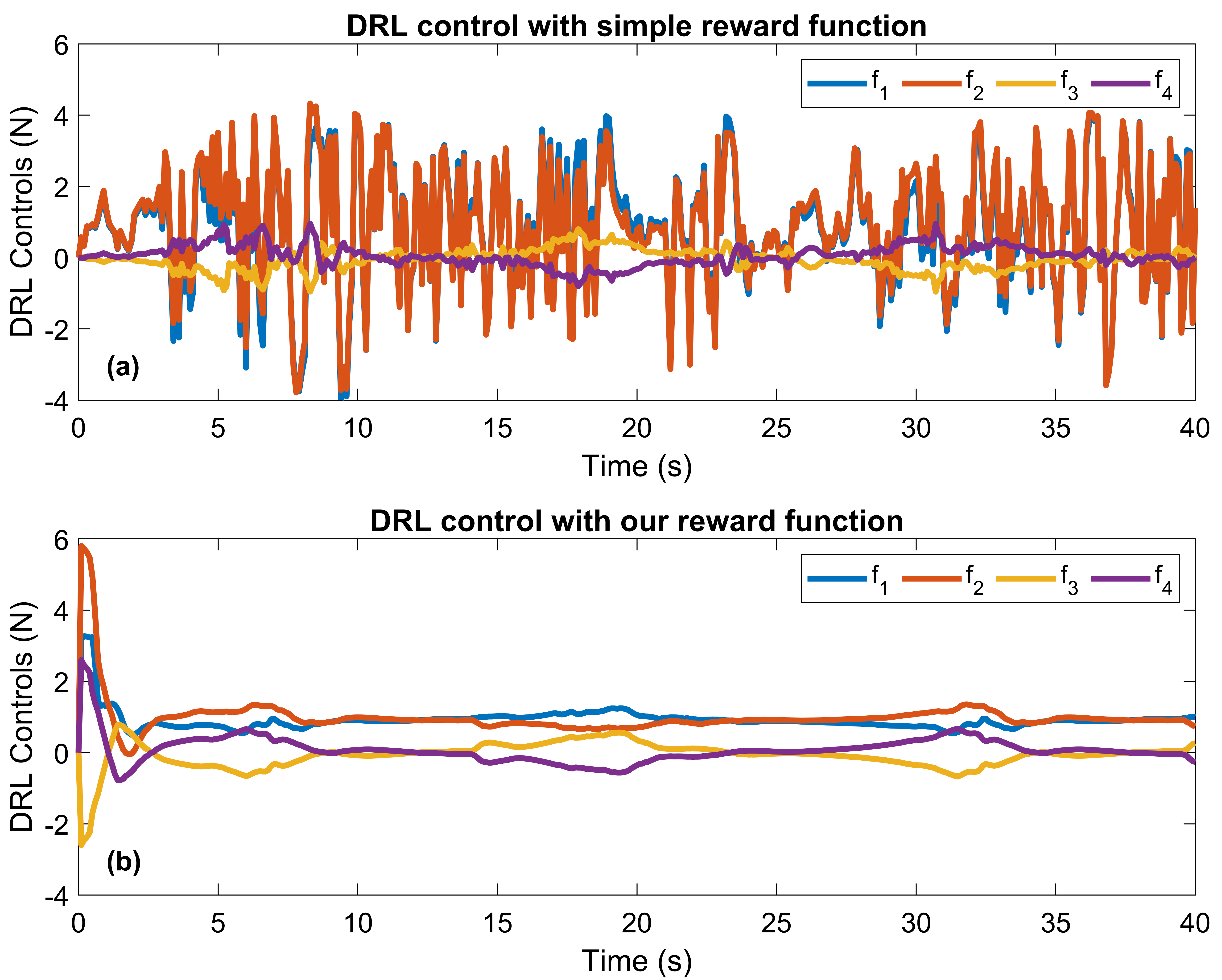}
        \caption{Control forces during DRL tracking using the simple reward function and our reward function.}
        \label{DRLbadrewardvsDRLreward-Force}
\end{figure}

In particular, we compared DRL control tracking performance with two different reward functions. One policy uses a simple reward function that addresses the terms (\ref{r_L}) and (\ref{e_{psi}}), while the other policy uses the developed reward function (\ref{r_{total}}) and contains five terms. As shown in Fig. \ref{DRLbadrewardvsDRLreward}, it is clear that the control policy proposed with our developed reward function goes beyond the control policy with the simple reward function. 
The mean tracking errors of both control policies for Euclidean distances are 0.0496 m and 0.0743 m, respectively, which means that the tracking errors of our control policy are 33.03\% lower compared to a policy utilizing only simple reward functions.
Fig. \ref{DRLbadrewardvsDRLreward-Force} compares the trajectory tracking control forces of the two policies.
Our control policy reduces energy consumption by 37.07\% and provides smooth actuation signals.

\subsubsection{Controller Comparison With Realistic Disturbances}

\begin{figure}[htb]
        \centering
        \includegraphics[width=1.0\linewidth] {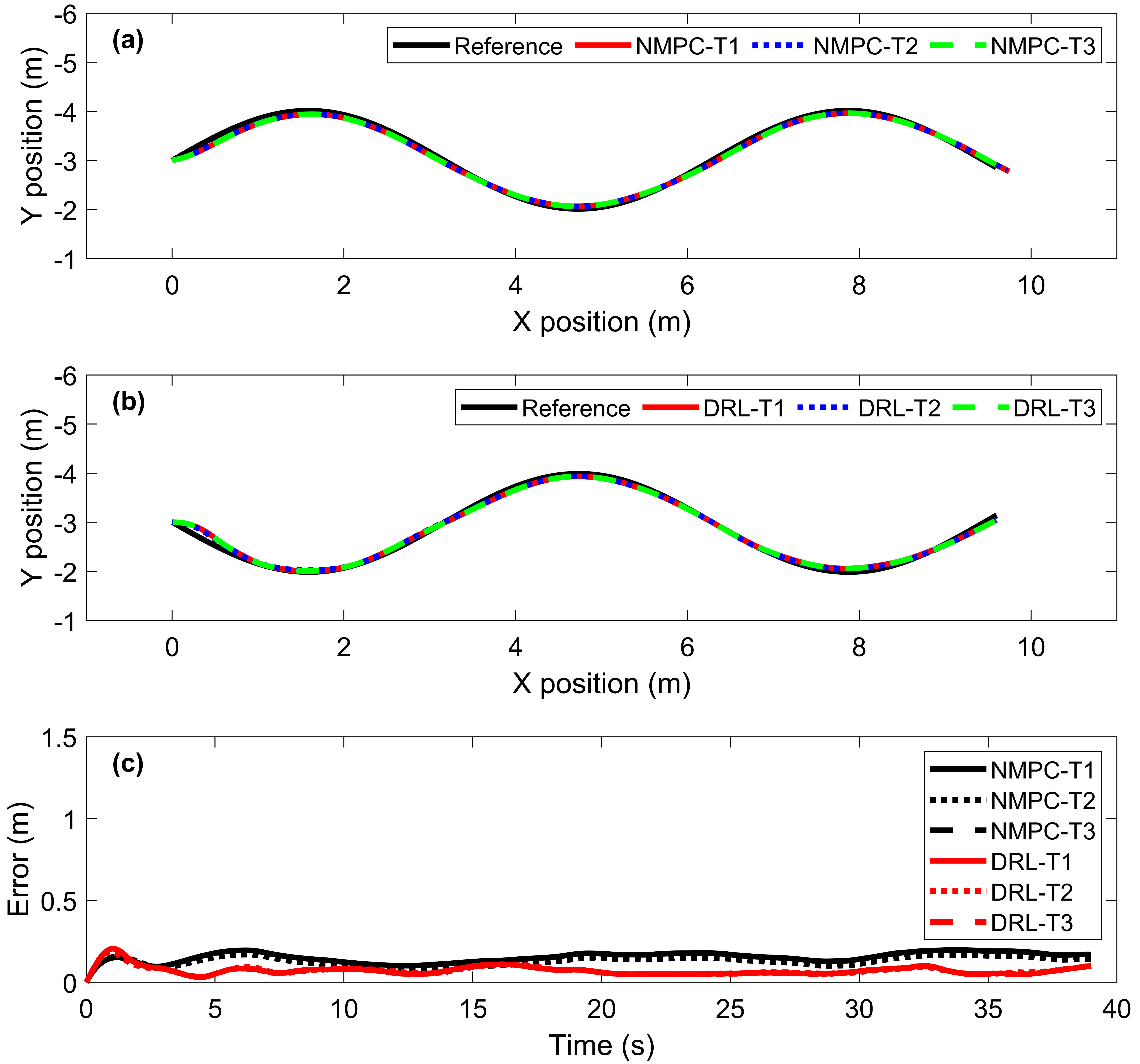}
        \caption{Performance of sinusoidal tracking using DRL and NMPC. (a) Trajectory and (b) tracking error. T1, T2, and T3 represent first, second and third tests, respectively.}
        \label{DRLvsMPCTrackingtrajectorSimWDist}
\end{figure}

We further compare the trained policy $\pi_{\theta}^{\ast}(\cdot)$ with the NMPC in \cite{WeiICRA2018}. 
We tested the controllers with different reference trajectories (C-Curve, Linear, and Sinusoidal) and both controllers can track the references well. Due to the limited space, the results of sinusoidal tracking are merely illustrated herein.
Figure \ref{DRLvsMPCTrackingtrajectorSimWDist} compares sinusoidal tracking between DRL and NMPC, considering model-based disturbances from wind, waves, and currents as $\bm{\tau}_{\text{env}} = \bm{\tau}_{\text{wind}} + \bm{\tau}_{\text{waves}} + (\mathbf{C}(\mathbf{v}_c)+\mathbf{D}(\mathbf{v}_c))\mathbf{v}_c$. Here, $\pmb{\tau}_{\text{wind}}$ represents the wind originated forces and moments, $\pmb{\tau}_{\text{wave}}$ stands for the wave induced forces and moments, and $\mathbf{v}_c$ the ocean current velocity. The disturbances dynamics are represented as follows:

\noindent \textbf{Wind}:
 Wind forces and moments depend on the wind speed $V_w$ and its angle of attack $\gamma_{rw}$, described by
\begin{equation}
\pmb{\tau}_{\text{wind}}
=
\begin{bmatrix}
	X_{\text{wind}}	\\
	Y_{\text{wind}} 	\\
	N_{\text{wind}} 
\end{bmatrix}
=
\dfrac{1}{2}\rho_a V_{rw}^2
\begin{bmatrix}
	C_X(\gamma_{rw})A_{FW}	\\
	C_Y(\gamma_{rw})A_{LW}	\\
	C_N(\gamma_{rw})A_{LW}L_{OA}
\end{bmatrix}
\end{equation}
where $V_{rw}$ is the wind relative speed, which can be computed from the wind velocity vector $\mathbf{v}_w = [V_w\cos(\beta_{\text{wind}}-\psi), V_w\sin(\beta_{\text{wind}}-\psi), 0]^T$, $\gamma_{rw} = -\text{atan2}(v_{rw},u_{rw})$, $\beta_{\text{wind}}$ stands for the wind direction, $\rho_a$ denotes the air density, $A_{FW}=0.045$ is the projected frontal area, $A_{LW}=0.09$ is the projected lateral area, and $L_{OA}=0.9$ is the ASV length. See \cite{GONZALEZGARCIA2021202} for more details on the computation of wind coefficients $C_X, C_Y, C_N$.

\noindent \textbf{Waves:}
Waves can be described combining first-order (zero-mean oscillatory motion) and second-order (wave drift) dynamics \cite{Fossen}. Then, the state-space equation follows an approximation of the Pierson-Moskowitz spectrum \cite{Fossen}, where $F_{\text{wave}}$ is the wave force, and $N_{\text{wave}}$ is the wave moment. The equations depend on Gaussian noise signals (to introduce randomness into the system), the encountered wave spectrum peak frequency, and the wave direction. For more details regarding the state-space representation, see \cite{Fossen,GONZALEZGARCIA2021202}. Finally, the wave disturbance vector is

\begin{equation}
\pmb{\tau}_{\text{wave}}
=
\begin{bmatrix}
	X_{\text{wave}}	\\
	Y_{\text{wave}} 	\\
	N_{\text{wave}} 
\end{bmatrix}
=
\begin{bmatrix}
	F_{\text{wave}}\cos(\beta_{\text{wave}} - \psi)	\\
	F_{\text{wave}}\sin(\beta_{\text{wave}} - \psi)	\\
	N_{\text{wave}}
\end{bmatrix}
\end{equation}

\noindent \textbf{Ocean Currents:} Currents are usually simplified to be irrotational, and modeled by a Gauss-Markov process \cite{Fossen}, such as
\begin{equation}
    \dot{V}_c + \mu_c V_c = \omega_c
\end{equation}
where $V_c$ denotes the magnitude of the current speed at direction $\beta_c$, $\omega_c$ is a Gaussian white noise signal, and $\mu_c >0$ is a constant. Then, the current velocity vector is $\mathbf{v}_c = [V_c\cos(\beta_c-\psi), V_c\sin(\beta_c-\psi), 0]^T$.

\begin{figure}[htb]
        \centering
        \includegraphics[width=0.9\linewidth] {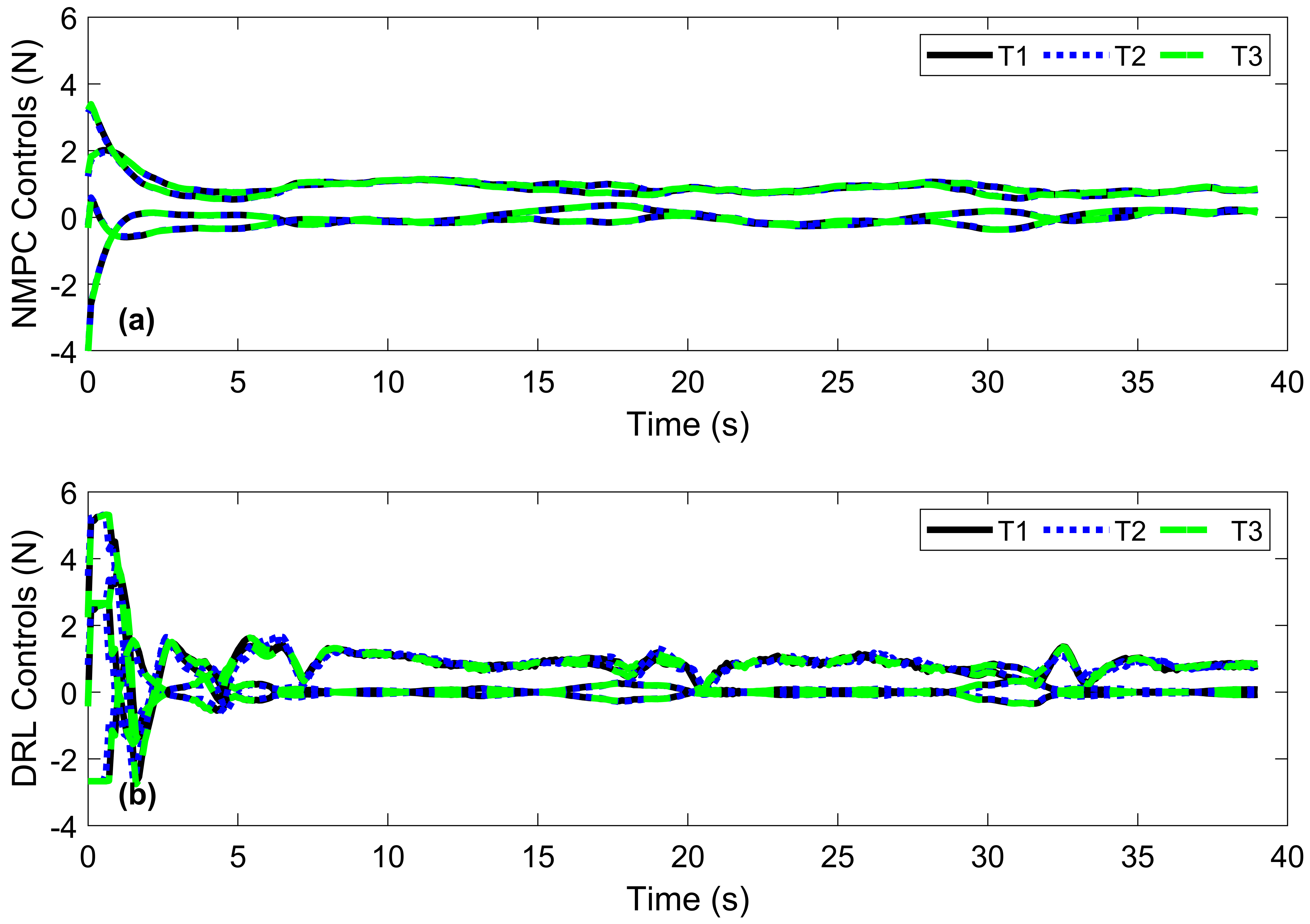}
        \caption{Control force while tracking the sinusoidal trajectory. (a) Using NMPC controller; (b) using DRL controller.}
        \label{DRLvsMPCControlsSimWDist}
\end{figure}

For this study, a wind speed of 4 knots was applied (a light breeze on the Beaufort scale \cite{Fossen}), while the waves provided forces of up to 1 N, and the currents a disturbance of up to 0.2 m/s. These magnitudes are enough to require around 25\% more control effort. The simulations for DRL and NMPC were each performed three times, ensuring repeatability. The Gaussian noises for waves and currents varied between simulation runs, providing different random perturbations around the same magnitudes.
 It can be observed that both the DRL and NMPC controllers can track well along the reference trajectory in simulation. The average tracking errors of the DRL controller and the NMPC controller in Euclidean distance are 0.068 m and 0.146 m,  which means that the tracking error for the DRL is 53.33\% less compared to that for the NMPC. Hence, NMPC has a larger error than the DRL in simulation, considering environmental disturbances.
 Moreover, Fig. \ref{DRLvsMPCControlsSimWDist} shows the control forces of DRL and NMPC. 
The average energy consumption for the NMPC and DRL are  2.018 N and 1.917 N respectively. 
Then, not only is the DRL more accurate than NMPC, but it also requires less energy to track the trajectory and reject the perturbations. Additionally, it is very interesting that the control forces behavior is similar between DRL and NMPC. This indicates that the DRL policy has learned a control performance similar to that of the NMPC controller, with enhanced robustness against external disturbances. 

\subsection{Experiments}
We deployed the trained control policy onto our surface vessel to track reference trajectories in the Charles River. We also used NMPC \cite{WeiICRA2018} as the benchmark controller and did the same tracking experiments as for the DRL. 
Both the DRL controller and the NMPC are run locally on the computer onboard the vessel. The execution time of the NMPC is approximately 1 ms while the execution time of the DRL is nearly 100 us. Moreover, the vessel runs a real-time SLAM \cite{TixiaoIROS2020a, WeiIROS2020a} using its onboard 3D LiDAR and IMU to estimate its pose $\bm{\eta}$ and velocity $\mathbf{v}$. 
All related parameters, including position, heading angle, velocity, and force of the robot, are recorded at 10 Hz during the experiments for analysis. 
During the field test, the wind on the Charles River was around 18 kph and the waves were less than one foot.

\begin{figure}[htb]
        \centering
        \includegraphics[width=0.9\linewidth] {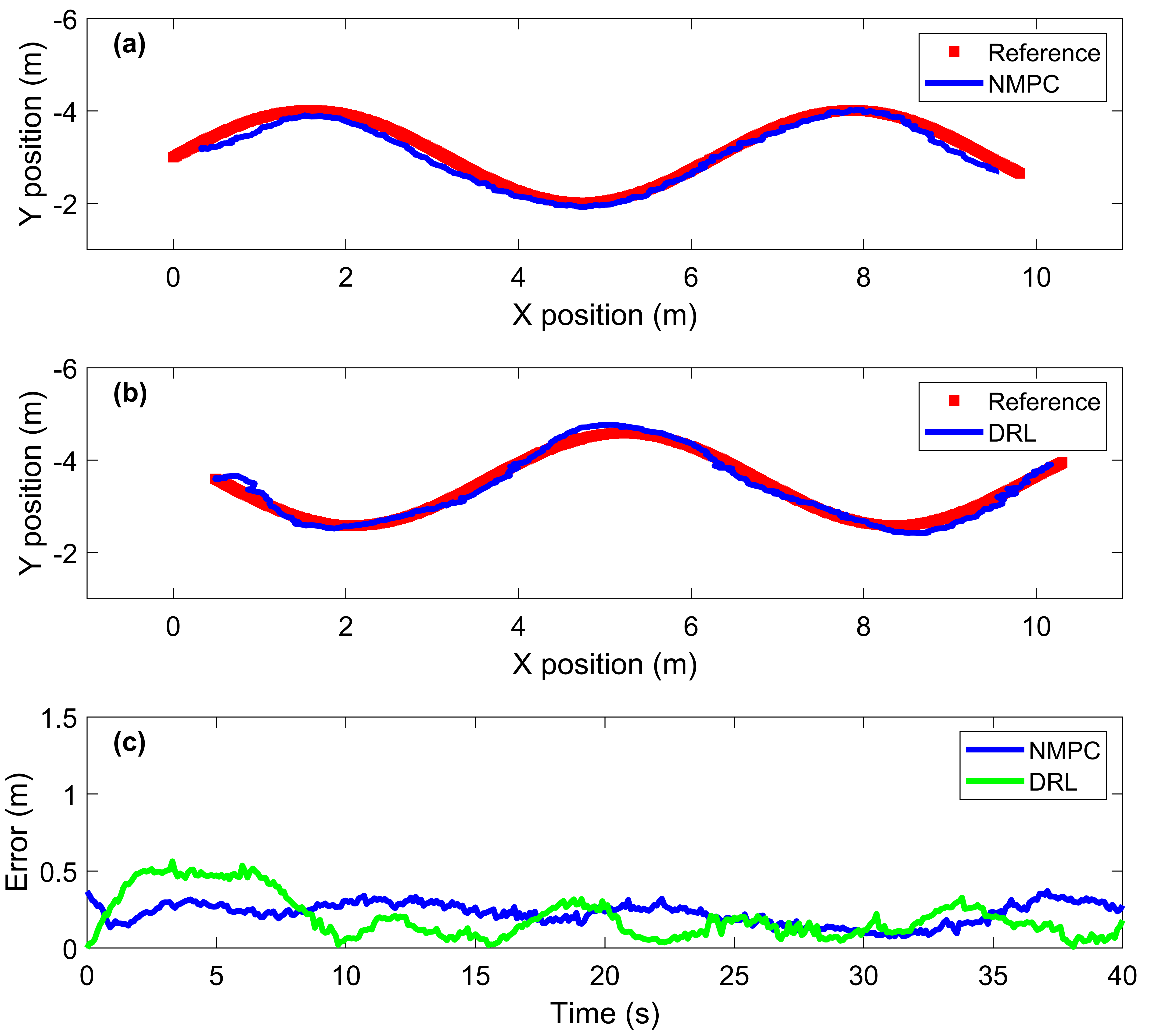}
        \caption{Performance of sinusoidal trajectory tracking in the Charles  River using DRL and MPC. (a) Trajectory and (b) tracking errors.}
        \label{DRLvsMPCTrackingtrajectoryRiver}
\end{figure}

\begin{figure}[htb]
        \centering
        \includegraphics[width=0.9\linewidth] {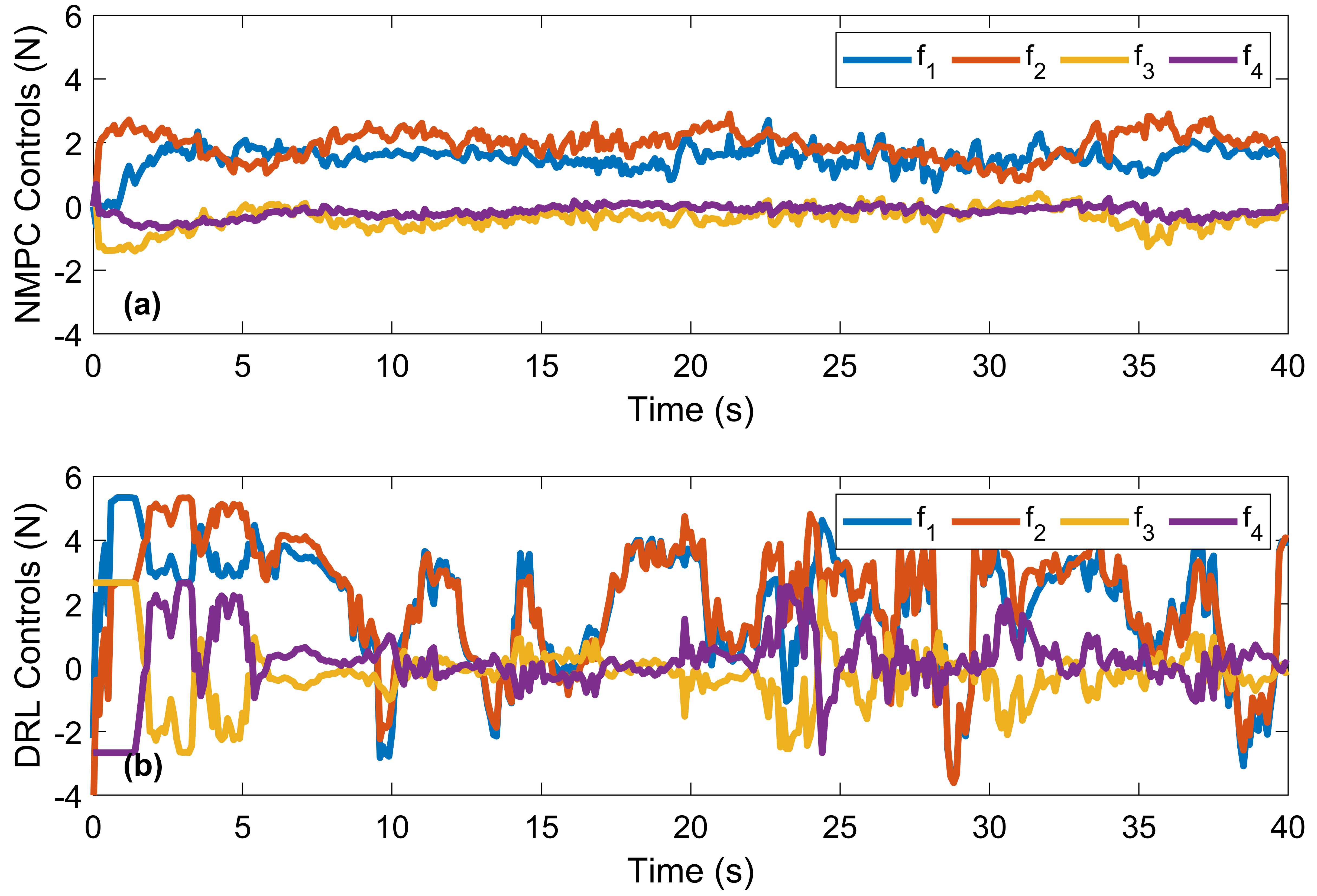}
        \caption{Control forces for the DRL controller and the MPC while tracking a sinusoidal trajectory in the Charles River. (a) NMPC (b) DRL.}
        \label{DRLvsMPCControlsRiver}
\end{figure}

Figure \ref{DRLvsMPCTrackingtrajectoryRiver} compared the trajectory tracking performance of the DRL and NMPC in the river with the aforementioned environmental disturbances. 
Tracking errors for the NMPC and the DRL are 0.2187 m and 0.1411 m, respectively, which means that the tracking error for the DRL is 35.51\% less compared to that for the NMPC. 
This suggests that the DRL controller has a better disturbance rejection ability than the NMPC.
Furthermore, the control forces for the DRL and NMPC while tracking the reference are illustrated in Fig. \ref{DRLvsMPCControlsRiver}.
In contrast to the simulations, it is clear that the control forces of the NMPC are smoother than those of the DRL controller. Likewise, the average energy consumption of the NMPC and the DRL is 3.978 N and 5.414 N, respectively. 
Then, the higher energy consumption and the larger variation within the actuator signals of the DRL is a consequence of its enhanced robustness ability.

\section{Conclusion and Future Work}
In this paper, we propose a DRL approach for controlling the trajectory tracking of an ASV. A differentiable non-linear dynamic model is used to construct an efficient simulation environment. We use a deep DRL approach to train the tracking control policy in a simulation environment with a series of sinusoidal curves as a reference trajectory.
To account for environmental disturbances, noisy measurements, and non-ideal actuators, multiple effective reward functions were delicately designed for the DRL tracking control policy. 
Numerical simulations subject to model-based environmental disturbances, and experiments in natural water show that the DRL policy can accurately track trajectories. In addition, in comparison to a state-of-the-art NMPC approach, the DRL performance presents a lower tracking error. In the experimental evaluation, DRL presents higher energy usage, a trade-off to counteract the external perturbations and achieve higher accuracy.
This work will be extended in two directions. First, we will use the DRL approach to obtain a more accurate dynamic model for our ASV, which can improve the fidelity of the simulation environment and, thereby, further improve the tracking capabilities. Second, we will consider combining DRL and MPC into one control framework so that we might benefit from both the higher tracking accuracy and the low energy consumption.

\clearpage 

\bibliographystyle{IEEEtran}
\bibliography{DRLControl}

\end{document}